\pdfoutput=1

\documentclass[11pt]{article}

\usepackage[]{acl}

\usepackage{times}
\usepackage{latexsym}

\usepackage[T1]{fontenc}

\usepackage[utf8]{inputenc}

\usepackage{microtype}
\usepackage{arydshln}
\usepackage{makecell}

\usepackage{hyperref}
\usepackage{url}
\usepackage{graphicx}
\usepackage{xspace,mfirstuc,tabulary}
\usepackage{times}
\usepackage{latexsym}
\usepackage{booktabs}
\usepackage{multirow}
\usepackage{comment}
\usepackage{dirtytalk}
\usepackage{color, colortbl}

\newcommand*{\yoruba}{Yor\`ub\'a\xspace}

\newcommand*\samethanks[1][\value{footnote}]{\footnotemark[#1]}

\title{$\varepsilon$ k{\' u} <mask>: Integrating {\yoruba} cultural greetings into machine translation}

\author{Idris Akinade\textsuperscript{1}\thanks{ \ \ Equal contribution.}, Jesujoba O. Alabi\textsuperscript{2}\samethanks, David Ifeoluwa  Adelani\textsuperscript{3},  \\
\textbf{Clement Odoje\textsuperscript{1}, and Dietrich Klakow\textsuperscript{2}} \\
\textsuperscript{1} Department of Linguistics and African Languages, University of Ibadan, Nigeria \\
\textsuperscript{2} Spoken Language Systems, Saarland University,
Saarland Informatics Campus, Germany\\
\textsuperscript{3} University College London, United Kingdom \\
\texttt{\{akinadeidris,lekeclement2\}@gmail.com} \\
\texttt{d.adelani@ucl.ac.uk, \{jalabi,dklakow\}@lsv.uni-saarland.de}
}

\begin{document}
\maketitle

\begin{abstract}
This paper investigates the performance of massively multilingual neural machine translation (NMT) systems in translating \yoruba greetings ($\varepsilon$ k{\' u} <mask>\footnote{For simplicity of notation in the title, we make use of $\varepsilon$ -- the Beninese \yoruba letter representation of \d {E} (which is used in Nigeria), and <mask> provides the context of greeting.}), which are a big part of \yoruba language and culture, into English. To evaluate these models, we present Ikini\yoruba, a \yoruba-English translation dataset containing some \yoruba greetings, and sample use cases. We analysed the performance of different multilingual NMT systems including Google Translate and NLLB and show that these models struggle to accurately translate \yoruba greetings into English. In addition, we trained a \yoruba-English model by finetuning an existing NMT model on the training split of Ikini\yoruba and this achieved better performance when compared to the pre-trained multilingual NMT models, although they were trained on a large volume of data. 

\end{abstract}

\section{Introduction}
In recent years, multilingual neural machine translation (NMT) models have shown remarkable improvement in translating both high and low-resource languages and have become widely used in various applications~\citep{kudugunta-etal-2019-investigating, aharoni-etal-2019-massively, nllb2022, bapna-51503}. Despite this progress, NMT models still struggle to accurately translate idiomatic expressions~\citep{fadaee-etal-2018-examining, baziotis-automatic}, cultural concepts such as proverbs~\citep{HAlkhresheh2018EnglishPI, adelani-etal-2021-effect}, and common greetings, particularly in African languages like \yoruba  -- a west African language, which has a rich cultural heritage.

Table \ref{tab:tab1} illustrates a \yoruba sentence containing frequently used greeting phrases by the \yoruba people, and the corresponding translations generated from three multilingual NMT systems, which are: Meta's NLLB~\citep{nllb2022}, 
Google Translate\footnote{\url{https://translate.google.com/} evaluated on 23rd January 2023}, and our own model.


\begin{table}[t]
\vspace{2mm}
    \centering\small
    \begin{tabular}{l}
    \toprule
    \textbf{Source:}  \textcolor{blue}{{\d E} k{\' u} oj{\' u}m{\d {\' o}}}, \textcolor{blue}{{\d e} s{\` i} k{\' u}} d{\' e}{\' e}d{\' e} {\` a}s{\` i}k{\` o} y{\` i}{\' i}. \\
    \textbf{Target:} \textcolor{blue}{Good morning} and \textcolor{blue}{compliment} for this period. \\
        \midrule
    \textbf{NLLB:} You have \textcolor{magenta}{died}, and you have \textcolor{magenta}{died} to this hour. \\
    \textbf{Google Translate:} \textcolor{magenta}{Die} every day, and \textcolor{magenta}{die} at this time. \\
    \textbf{Our Model:} \textcolor{blue}{Good morning} and \textcolor{blue}{compliment} for this time. \\
    \bottomrule
    \end{tabular}
    \caption{Translation outputs of 3 different NMT models.}
    \label{tab:tab1}
\end{table}

An examination of NLLB and Google Translate's model outputs reveals that they all fail to produce accurate translations for the input sentence. One possible explanation for this is the lack of sufficient training data including these types of greetings, even though they were trained on a large volume of multilingual data. Furthermore, \textit{k{\' u}}, a common word in these kinds of greetings, has two main interpretations that could mean either death or a compliment, depending on the context. Similarly, the syntactic frame of occurrence also determines the meaning of the verb (the type of complement and adjunct), and this is due to the ambiguous nature of \yoruba verbs. Hence, it is possible that these models were trained on data with \textit{k{\' u}} having the meaning death. 


To address this issue, this paper introduces a new  dataset dubbed Ikini\yoruba, a \yoruba-English translation dataset of popular \yoruba greetings. We evaluate the performance of existing multilingual NMT systems on this dataset, and the results demonstrate that although current multilingual NMT systems are good at translating \yoruba sentences into English, they struggle to accurately translate \yoruba greetings, highlighting the need for further research in translating such cultural concepts on low-resource African languages. 

\section{\yoruba cultural greetings}
\label{sec:sec2p1}

\yoruba is a language spoken by the \yoruba people. It is native to Nigeria, Benin and Togo with an estimate of over 40 million speakers~\cite{ethnologue}. \yoruba makes use of 25 Latin letters excluding the Latin characters  (c, q, v, x and z), and additional letters ({\d e}, gb, {\d s} , {\d o}).  \yoruba is a tonal language with three tones: low, middle and high. These tones are represented by the grave (e.g. ``\`{a}
''), optional macron (e.g. ``\={a}'') and acute (e.g. ``\'{a}'') accents respectively.

Greetings are inseparable from the \yoruba people since they are important for first impressions and are even considered to be a part of \yoruba identity. After the abolition of the slave trade at the beginning of the 19th century, the \yoruba indigenes who were rescued by the British warship settled in Freetown, a place in present-day Sierra Leone. People began to call them \textit{{a} k{\' u}} which is a fragment attached to all forms of greetings in \yoruba~\citep{webster_1966}. This is because while an English speaker 
will say \textit{good morning}, \textit{happy birthday}, \textit{merry Christmas}, and so on, the \yoruba people would say \textit{{\d e} k{\' a}{\` a}r{\d{\` o}}}, \textit{{\d e} k{\' u} {\d o}j{\d{\' o}} {\` i}b{\' i}}, and \textit{{\d e} k{\' u} {\d o}d{\' u}n k{\' e}r{\' e}s{\` i}mes{\` i}}.
The recurrence of \textit{{\d e} k{\' u}} in their everyday conversation resulted the appellation \textit{a k{\' u}}. 

\textit{{\d E} k{\' u}} has the same semantic importance as `good-', `merry-' and `happy-' in English greetings. Without the fragment \textit{{\d e} k{\' u}} in the communication frame of greeting, the cultural knowledge shared by interlocutors will be lost.  

Structurally, \textit{{\d e} k{\' u}} can be syntactically explained to have a subject-predicate relationship, rather than being a single lexeme or a prefix as claimed by most scholars. Using the paradigmatic relationship \citep{saussure1916course,asher1994encyclopedia} lens, \textit{{\d e}} can be replaced with any pronoun or nominal item (as described by interlocutors) with +human feature and still fit in perfectly. The +human feature is necessary because compliments are mainly for humans and \textit{k{\' u}} requires a selection restriction to sieve out the non-human elements. Table \ref{tab:tab2} shows some of these constructions. It is equally important to note here that \textit{{\d e} k{\' u}} can also be used for supernatural beings or metaphysical beings which in this form sounds like a personification. 

\begin{table}[t]
    \centering\small
    \begin{tabular}{l p{2.2cm}p{2.1cm}}
    \toprule
    Greeting & Person & Meaning \\
    \midrule
    \textit{O k{\' u} {\` i}r{\` i}n} & 2nd person singular & Compliment for walking \\
    \textit{{A} k{\' u} {\` o}d{e}} & 1st person plural & Compliment for attending a party \\
    \textit{W{\d{\' o}}n k{\' u} {\` i}j{\' o}{\` o}k{\' o}} & 3rd person plural & Compliment for sitting \\
    \bottomrule
    \end{tabular}
    \caption{Some \textit{{\d E} k{\' u}} constructions }
    \label{tab:tab2}
\end{table}

\textit{K{\' u}} on the other hand is a transitive predicate that requires a compliment. This compliment could either be a noun that signifies time like \textit{{\` a}{\' a}r{\d{\` o}}} (morning), a noun that denotes season like \textit{{\d {\` o}}ririn}/\textit{{\` o}t{\' u}t{\` u}} (cold), a noun that points to a celebration like \textit{k{\' e}r{\' e}s{\` i}mes{\` i}} (Christmas), a nominalized verb that describes an event or action like \textit{{\` i}j{\' o}k{\` o}{\' o}} (sitting), and many more. Omitting the compliment in a greeting construction will alter the interpretation of the expression which may also change the meaning of \textit{k{\' u}} to death.   

\section{Related Work}
The development of machine translation systems for low-resource languages such as \yoruba has seen a significant amount of research efforts in recent years. One major area of focus has been on curating translation datasets for these languages, which are collected using either automatic or manual methods. Examples of automatically collected datasets that include \yoruba are JW300~\citep{agic-vulic-2019-jw300}, CCMatrix~\citep{schwenk-etal-2021-ccmatrix}, and CCAligned~\citep{el-kishky-etal-2020-ccaligned}. On the other hand, examples of manually translated datasets for Yoruba include MENYO-20k~\citep{adelani-etal-2021-effect}, MAFAND-MT~\citep{adelani-etal-2022-thousand}, FLORES-101~\citep{goyal-etal-2022-flores}, and NTREX~\citep{federmann-etal-2022-ntrex}. These datasets have been instrumental in the study, development, and improvement of machine translation systems for \yoruba. 

For example,~\citet{adelani-etal-2021-effect} investigated how domain data quality and the use of diacritics, a crucial aspect of \yoruba orthography, impact \yoruba-English translations. \citet{adebara-etal-2022-linguistically} examined the effectiveness of \yoruba-English machine translation in translating bare nouns (BN), by comparing the results obtained from using statistical machine translation methods and neural approaches.~\citet{adelani-etal-2022-thousand} investigated how to effectively leverage pre-trained  models for translation of African languages including \yoruba. Despite the attempts to create datasets and develop translation systems for \yoruba, to the best of our knowledge, only~\citet{adelani-etal-2021-effect} has examined a cultural aspect of \yoruba by evaluating their models on \yoruba proverbs, which are a significant part of \yoruba tradition. However, this research has not looked into how these models perform on another cultural aspect which is \yoruba greetings. Furthermore, there appear to be no prior works that have evaluated machine translation performance specifically for this aspect of the language and for other languages. Therefore, in this work, we investigate the performance of \yoruba-English translation models on \yoruba greetings.

\section{Ikini\yoruba corpus}

\paragraph{Greetings dataset:} We introduce \textbf{Ikini\yoruba}, a \yoruba-English translation dataset for \yoruba greetings and their usage in various contexts, containing $960$ parallel instances. The data curation process involved three key stages. Firstly, we gathered commonly used \yoruba greetings that cover a variety of situations such as time, season, celebration, and more, as outlined in Section \ref{sec:sec2p1}, resulting in a total of $160$ \yoruba greetings. Secondly, we created $5$ different example sentences for each greeting, where the greetings are used in context, by native speakers of the language, resulting in $800$ use cases in total. Lastly, we asked an expert translator to translate the seed data and the use cases into English. We split the created data into train/dev/test splits with $100$/$20$/$40$ seed greeting instances. For each instance in a split, the $5$ example sentences created are assigned to the same split.   

\paragraph{Conversational dataset:} For our experiments, we used the movie transcripts subset of the MENYO-20k~\cite{Adelani2020MENYO20kAM} dataset, which is a human-translated English-\yoruba dataset for movie transcripts. We selected this dataset because it consists of conversational data.

Table \ref{tab:data_source1} shows the sample sentences in the Ikini\yoruba dataset and Movie Transcript datasets, while Table \ref{tab:data_source2} highlights the statistics of these datasets.

\begin{table}[t]
\begin{center}
\resizebox{\columnwidth}{!}{%
\begin{tabular}{p{40mm} p{40mm}}
\toprule
\textbf{\yoruba} & \textbf{English}   \\
\midrule
\multicolumn{2}{l}{\textbf{Ikini\yoruba - Seed Greetings}} \\
\addlinespace[0.5em]
{\d E} k{\' u} {\` i}f{\d {\' e}} & Thanks for the love \\
{\d O}k{\d {\` o}} {\' a} r{\` e}f{\` o}  & Safe ride \\
\addlinespace[0.5em]
\multicolumn{2}{l}{\textbf{Ikini\yoruba - Greetings with contexts}} \\
\addlinespace[0.5em]
{\d E} k{\' u} {\` i}f{\d {\' e}}, Ire l{\` a} {\' o} m{\' a} b{\' a} ara wa {\d s}e. & Thanks for the love, may we continue to celebrate one another.  \\
A {\' o} ma foj{\' u} s{\' o}n{\` a} l{\' a}ti r{\' i}i y{\' i}n, {\d o}k{\d {\` o}} {\' a} r{\` e}f{\` o} & Looking forward to seeing you, safe ride. \\
\addlinespace[0.5em]
\midrule
\addlinespace[0.5em]
\multicolumn{2}{l}{\textbf{Movie Transcript}} \\
\addlinespace[0.5em]
{\d E} k{\' a}{\` a}s{\' a}n ma. & Good afternoon ma.	\\
{\d E} \`{n}l{\d {\` e}} {\d s}{\` a}! Mo m{\d {\` o}} y{\' i}n & Hello sir! I know you \\
F{\d {\' e}}mi k{\' i} l{\' o} {\d s}{\d e}l{\d {\` e}} b{\' a}y{\` i}{\' i}? & Femi what is it now?	\\
Gbogbo nnk{\` a}n {\' a} d{\' a}ra, a j{\d o} w{\` a} n{\' i}n{\' u} {\d {\` e}} ni & Everything will be fine, we're in this together \\
\bottomrule
\end{tabular}
}
\end{center}
\footnotesize
  \caption{Sample sentence pairs from the Ikini\yoruba and the Movie Transcripts datasets.}
     \label{tab:data_source1}
\end{table}


\section{Experiments}

\subsection{Experimental Setup}
Greetings play a crucial role in \yoruba culture and are widely used in daily conversations by \yoruba people. For every action, there is a customary way of greeting or complimenting those involved using the phrase \textit{\d{E} k{\' u}}. 
In this work, we compare several existing translation systems and evaluate their performance on \yoruba greetings. We demonstrate the effectiveness of these translation systems by testing them on movie transcripts, which are conversational in nature. Below, we outline our experiments.

\begin{table}[t]
\footnotesize
\begin{center}
\begin{tabular}{p{15mm}rrr}
\toprule
 &  \multicolumn{3}{c}{\textbf{Number of Sentences}}  \\
\textbf{Data}  & \textbf{train} & \textbf{dev} & \textbf{test}  \\
\midrule
    \textit{IkiniYoruba} & $600$ & $120$ & $240$ \\
    \textit{Movie Transcript} & -- & -- & $775$ \\
\bottomrule
\end{tabular}
\end{center}
\footnotesize
  \caption{The split of the data}
     \label{tab:data_source2}
\end{table}

\paragraph{Translation Models:} In this study, we evaluate the performance of three multilingual NMT systems. These systems were pre-trained on various languages, and they are Google multilingual NMT, the distilled version of Meta's NLLB~\citep{nllb2022} with 600M parameters, and a publicly available M2M-100~\citep{fan2020beyond} with 418M parameters fine-tuned on the MENYO-20k dataset. We generated translations for the test sets using the Google Translate web application\footnote{\url{https://translate.google.com/} evaluated on 23rd January 2023}, while for Meta's M2M-100 and NLLB models, we used the HuggingFace transformers\footnote{\url{https://github.com/huggingface/transformers}} library. 

\paragraph{Data preprocessing and evaluation:} To standardize the format of the two parallel datasets, we converted the \yoruba texts in the dataset to Unicode Normalization Form Composition (NFC). And to automatically assess the performance of the models, we used BLEU~\citep{papineni-etal-2002-bleu} score implemented in SacreBLEU\footnote{\texttt{case:mixed|eff:no|} \texttt{tok:13a|smooth:exp|version:2.3.1}}~\citep{post-2018-call}. 

\begin{table*}[ht]
\footnotesize
\begin{center}
\begin{tabular}{l|cc|cc}
\toprule
\multicolumn{5}{c}{\textbf{yo $\rightarrow$ en}} \\
& \multicolumn{2}{c}{\textbf{BLEU}} & \textbf{Adequacy} & \textbf{CCP} \\
& \textbf{Movie Transcript} & \textbf{Ikini\yoruba} & \multicolumn{2}{c}{\textbf{Ikini\yoruba}} \\
\midrule
Google Translate & $31.05$ & $9.47$ & $2.02$ & $0.11$ \\
NLLB & $27.19$ & $5.03$ & $1.88$ & $0.09$ \\
M2M-100 & $34.70$ & $4.33$ & $1.73$ & $0.05$  \\
{ } + Ikini\yoruba & $26.05$ & $\textbf{29.67}$ & $\textbf{2.79}$ & $\textbf{0.35}$ \\
{ } + Movie Transcript & - & $6.25$ & - & - \\
{ } + Ikini\yoruba + Movies Transcript & - & $29.49$ & - & - \\
\bottomrule
\end{tabular} 
\end{center}
\footnotesize
  \caption{Performance of the models on Ikini\yoruba and Movie Transcript. The M2M-100 and NLLB models have 418M and 600M parameters respectively. CCP is Cultural Content Preservation and it indicates whether greetings/compliments within the source sentences are preserved or not in the translation outputs. }
     \label{tab:result}
\end{table*}

\begin{table}[t]
\footnotesize
\centering
\resizebox{\columnwidth}{!}{%
\begin{tabular}{p{2mm}lp{62mm}}
\toprule
1. & \textbf{Source} & A \textcolor{blue}{k{\' i}} {\` a}w{\d o}n k{\` i}r{\` i}s{\` i}t{\d {\' e}}ni \textcolor{blue}{k{\' u} {\d o}d{\' u}n {\` A}j{\' i}nde}. \\
& \textbf{Target} & We \textcolor{blue}{greet} the Christians a \textcolor{blue}{happy Easter}. \\
\addlinespace[0.2em]
\specialrule{0.1pt}{0.1pt}{0.1pt}
\addlinespace[0.2em]

& \textbf{Google T.} & We \textcolor{blue}{wish} Christians a \textcolor{blue}{happy Easter}. \\
& \textbf{NLLB} & \textcolor{blue}{Celebrations} are celebrated on \textcolor{magenta}{New Year's Eve}. \\
& \textbf{M2M-100} & We \textcolor{blue}{greeted} \textcolor{magenta}{ridiculers} in the \textcolor{blue}{resurrection} \textcolor{magenta}{year}. \\
& \textbf{Our Model} & We \textcolor{blue}{greet} the \textcolor{magenta}{hardworking people} the \textcolor{blue}{resurrection celebration}. \\
\bottomrule
\addlinespace[0.5em]

2. & \textbf{Source} & \textcolor{blue}{{\d E} k{\' u} {\` a}p{\` e}j{\d e}} {\d {\` e}}yin ol{\' o}y{\` e}. \\
& \textbf{Target} & \textcolor{blue}{Happy feasting} chiefs. \\
\addlinespace[0.2em]
\specialrule{0.1pt}{0.1pt}{0.1pt}
\addlinespace[0.2em]
& \textbf{Google T.} & \textcolor{magenta}{Farewell to the party}, you chiefs. \\
& \textbf{NLLB} & \textcolor{blue}{Enjoy the feast}, you leaders. \\
& \textbf{M2M-100} & You chieftains \textcolor{magenta}{die at the banquet}. \\
& \textbf{Our Model} & \textcolor{blue}{Compliment} for a \textcolor{magenta}{reception} chiefs. \\
\bottomrule
\addlinespace[0.5em]

3. & \textbf{Source} & \textcolor{blue}{{\d E} {\` n}l{\d {\' e}} o} {\d {\` e}}yin {\` e}{\` e}y{\` a}n mi, \textcolor{blue}{{\d s}e {\` a}l{\` a}{\' a}f{\' i}{\` a} ni?} \\
& \textbf{Target} & \textcolor{blue}{Hello} my people, \textcolor{blue}{I hope you are fine?} \\
\addlinespace[0.2em]
\specialrule{0.1pt}{0.1pt}{0.1pt}
\addlinespace[0.2em]
& \textbf{Google T.} & My people, \textcolor{blue}{is it peace?} \\
& \textbf{NLLB} & \textcolor{blue}{Is it peace}, my people? \\
& \textbf{M2M-100} & May you, my people, be at \textcolor{blue}{peace}? \\
& \textbf{Our Model} & \textcolor{blue}{Hello} my people, \textcolor{blue}{hope you are fine?} \\
\bottomrule
\addlinespace[0.5em]

4. & \textbf{Source} & \textcolor{blue}{O k{\' u} {\` a}j{\` a}b{\d {\' o}}} {\d {\` o}}r{\d {\' e}} mi. \\
& \textbf{Target} & \textcolor{blue}{Compliment for escaping danger} my friend.	 \\
\addlinespace[0.2em]
\specialrule{0.1pt}{0.1pt}{0.1pt}
\addlinespace[0.2em]
& \textbf{Google T.} & \textcolor{magenta}{You are dead} my friend. \\
& \textbf{NLLB} & \textcolor{magenta}{You sacrificed my friend}. \\
& \textbf{M2M-100} & \textcolor{magenta}{You lost my friend's womb}. \\
& \textbf{Our Model} & \textcolor{blue}{Compliment for escaping the danger} of my friend. \\
\bottomrule
\addlinespace[0.5em]

5. & \textbf{Source} & \textcolor{blue}{O k{\' u} ay{\d e}y{\d e} {\d o}j{\d {\' o}} {\` i}b{\' i}} Ol{\' u}wad{\' a}mil{\' a}re. \\
& \textbf{Target} & \textcolor{blue}{Happy birthday celebration} Ol{\' u}wad{\' a}mil{\' a}re. \\
\addlinespace[0.2em]
\specialrule{0.1pt}{0.1pt}{0.1pt}
\addlinespace[0.2em]

& \textbf{Google T.} & He \textcolor{magenta}{died celebrating the birthday} of the Almighty. \\
& \textbf{NLLB} & You \textcolor{magenta}{celebrated} the Righteous One's \textcolor{magenta}{birthday}. \\
& \textbf{M2M-100} & You \textcolor{magenta}{died on the anniversary of the birth} of Olúwádámiler. \\
& \textbf{Our Model} & Compliment for \textcolor{magenta}{today’s anniversary} of God's goodwill. \\
\bottomrule
\addlinespace[0.5em]
\end{tabular} 
}
  \caption{Examples of MT output for different NMT models. Examples selected from the test set. }
  \label{tab:discuss1}
\end{table}

\subsection{Experimental results}
Table \ref{tab:result} shows the results of evaluating the three different models on the two datasets: Ikini\yoruba test split and Movie Transcripts. The models obtained impressive performance on the Movie Transcript data with high BLEU scores but poorly on the Ikini\yoruba data with significantly lower scores. This highlights their inability to translate \yoruba cultural content such as greetings. The best-performing model, M2M-100, had a BLEU score of $34.70$ on Movie Transcript data as it was trained on this same data by its authors. However, it had a score of $4.3$ on greetings data. The second-best model, Google Translate, was $3.65$ points below the best model on Movie Transcript. It performed better on greetings data with a score of $9.47$, though still lower compared to its performance on Movie Transcript data.


In addition, we finetuned the M2M-100 model on Ikini\yoruba, Movie Transcripts, and a combination of both data sources and evaluated the models on the Ikini\yoruba test split. Our results show that finetuning the M2M-100 on Movie Transcripts improves the model's performance on Ikini\yoruba by $1.92$ BLEU points compared to the original M2M-100. However, the best performance was achieved when the M2M-100 was finetuned on the Ikini\yoruba training split, with a BLEU score of $29.67$. Finetuning the M2M-100 on the combination of both datasets did not result in any improvement. We do not evaluate the M2M-100 model finetuned on MovieTranscript data on the MovieTranscript data, as this would result in evaluating on the same data used for training. 

To understand the performance of individual models on the Ikini\yoruba test set, we conducted human evaluations of the translated outputs from Google Translate, NLLB, M2M-100, and M2M-100 finetuned on the Ikini\yoruba dataset. We asked three native \yoruba speakers fluent in English to rate the 240 sentences for each system on two criteria: adequacy (on a Likert scale of 1 to 5) and cultural content preservation - CCP (binary scale of 0 or 1). Here, adequacy describes how much of the meaning of the reference translation was preserved in the MT output, and CCP indicates whether the greetings/compliments within the translation are preserved or not.  The results show that the NMT systems struggle at translating \yoruba greetings accurately, and they confirm the results of the automatic evaluation, showing that M2M-100 finetuned on IkiniYorùbá outperforms all other models. Overall, we observed that human evaluation shows moderate agreement with automatic evaluation. 

\subsection{Qualitative analysis of translation outputs}
In Table \ref{tab:discuss1}, we present some translation outputs from the different models for 5 \yoruba sentences sampled from the Ikini\yoruba test split. 

Google Translate and NLLB perform well in some cases by generating translations that were similar and contextually appropriate, for instance, in the second and third examples. Google Translate gave the most similar output to the target sentence in the first example. Our model in this instance translated `{\d o}d{\' u}n' (meaning `year' in isolation or `celebration' when it occurs alone with {\d e} k{\' u}) quite independently `{\` a}j{\' i}nde' (meaning `resurrection' in isolation). Hence, `resurrection celebration' appears in the output. NLLB fails in this example but in the second example, it gives the closest contextual interpretation while our model got everything right except `{\` a}p{\` e}j{\d e}' which is translated as `reception' instead of `feasting'.

Our model outperforms Google Translate and NLLB in the third and fourth examples. It generated nearly identical output to the target sentence, thereby  showing the preservation of both cultural content and semantic interpretation ability learned from the training data. In contrast, both Google Translate and NLLB were unsuccessful in producing the correct translation. The third example is an inquiry about well-being and it is, therefore, appropriate to use the word `fine', and not `peace'. In the fourth example, our model also shows to have an understanding of the contextual usage of \textit{k{\' u}} as a compliment which both Google Translate and NLLB failed to do. In addition, similar to the automatic evaluation result, our model generated better outputs when compared to M2M-100 which was the base model on which it was trained, confirming the ability of the model to learn from a few training instances even for low-resource languages such as \yoruba~\cite{adelani-etal-2022-thousand}.

However, all the models failed in the last example. The models incorporated the concept of celebration or birthday in their output, but none of them were able to produce output that was exactly or semantically equivalent to the target sentence. A mistake common to all the model output except for M2M-100, is that they tried to translate `Ol{\' u}wad{\' a}mil{\' a}re'\footnote{translates to: {`the lord justifies me'}, but the models still failed in this case.} which is a name of a person and should not be translated. Hence, there is a need for more effort in solving this greetings translation task, either by creating more data or developing better approaches at translating these greetings into English. 

\section{Conclusion}
In this study, we analyzed the performance of machine translation models in translating \yoruba greetings into English. To achieve this objective, we introduced a novel dataset called Ikini\yoruba, which contains a collection of \yoruba greetings and their respective sentence use cases. We evaluated three publicly available machine translation models on this dataset and found that, despite their ability to translate other \yoruba texts, they failed to accurately translate \yoruba greetings, which are a crucial aspect of \yoruba culture. In future research, we aim to expand the Ikini\yoruba dataset by adding more profession-based greetings and exploring ways to enhance the performance of machine translation models with these data.

\section*{Limitations}
One of the main limitations of our study is the lack of parallel data for \yoruba greetings. Hence, we had to create Ikini\yoruba, which has 960 parallel sentences and may not be representative of all the greetings in \yoruba language including profession-based greetings. In addition, our study did not explore the use of verb disambiguation methods or external knowledge bases, to enhance the performance of our models. We leave these for future research.

\section*{Acknowledgements}
 We appreciate Dr. Ezekiel Soremekun for the initial discussion that led to this work. We are grateful for the  feedback from Dr. Rachel Bawden, Vagrant Gautam and anonymous reviews from AfricaNLP and C3NLP. Moreover, we would like to thank Timileyin Adewusi, Ganiyat Afolabi, and Oluwatosin Koya who took part in the human evaluation process. Jesujoba Alabi was partially funded by the BMBF project SLIK under the Federal Ministry of Education and Research grant 01IS22015C. David Adelani acknowledges the support of DeepMind Academic Fellowship programme. 

\bibliography{anthology,custom}
\bibliographystyle{acl_natbib}



\end{document}